\documentclass{new_tlp}

\usepackage[utf8]{inputenc}
\usepackage[T1]{fontenc}
\usepackage{amsmath,amssymb}
\usepackage{microtype}
\usepackage{listings}
\usepackage{pifont}
\newcommand{\head}[1]{\ensuremath{\mathit{H}(#1)}} % {\ensuremath{\mathit{head}(#1)}}
\newcommand{\sysfont}{\textit}

\newcommand{\aspartame}{\sysfont{aspartame}}

\newcommand{\aspmttosmt}{\sysfont{aspmt2smt}}

\newcommand{\assat}{\sysfont{assat}}

\newcommand{\clasp}{\sysfont{clasp}}
\newcommand{\clingcon}{\sysfont{clingcon}}
\newcommand{\clingo}{\sysfont{clingo}}
\newcommand{\cmodels}{\sysfont{cmodels}}

\newcommand{\cplex}{\sysfont{cplex}}
\newcommand{\dingo}{\sysfont{dingo}}
\newcommand{\dlvhex}{\sysfont{dlvhex}}

\newcommand{\ezcsp}{\sysfont{ezcsp}}
\newcommand{\ezsmt}{\sysfont{ezsmt}}

\newcommand{\gecode}{\sysfont{gecode}}

\newcommand{\inca}{\sysfont{inca}}

\newcommand{\lpsolve}{\sysfont{lpsolve}}

\newcommand{\mingo}{\sysfont{mingo}}

\newcommand{\wasp}{\sysfont{wasp}}

\newcommand{\zzz}{\sysfont{z3}}

\newcommand{\python}{Python}
\newcommand{\prolog}{Prolog}

\newcommand{\cpp}{C++}

\newcommand{\clingod}[1]{\clingo\textnormal{[}\textsc{#1}\textnormal{]}}
\newcommand{\ASPm}[1]{ASP\raisebox{.7pt}{[\textsc{#1}]}}
\newcommand{\T}{\textsc{t}}
\newcommand{\TO}{\textsc{to}}
\newcommand{\dl}{\textsc{dl}}
\newcommand{\lc}{\textsc{lc}}
\newcommand{\tsp}{\textsc{2sp}}
\newcommand{\fs}{\textsc{fs}}
\newcommand{\js}{\textsc{js}}
\newcommand{\os}{\textsc{os}}
\newcommand{\is}{\textsc{is}}
\newcommand{\rf}{\textsc{rf}}
\newcommand{\ws}{\textsc{ws}}
\newcommand{\class}{\textsc{class}}
\newcommand{\clingolc}[3]{\textsc{#1}/\textsc{#2}/\textsc{#3}}
\newcommand{\atomtype}[1]{\textsc{#1}}
\newcommand{\lcatype}[2]{#1+#2}
\newcommand{\na}{--}
\newcommand{\PS}[2]{\ensuremath{#1|_{\mathcal{#2}}}}
\newcommand{\smset}[1]{\ensuremath{\mathcal{X}(#1)}}
\newcommand{\smsetlc}[1]{\ensuremath{\mathcal{X}_{\textsc{lc}}(#1)}}

\newtheorem{theorem}{Proposition}

\newcommand{\cmark}{\ding{51}}%
\newcommand{\xmark}{\ding{55}}%
\usepackage{threeparttable}
\usepackage{booktabs}
\usepackage{url}

\lstdefinelanguage{clingo}{
  keywordstyle=[1]\usefont{OT1}{cmtt}{m}{n},%
  keywordstyle=[2]\textbf,%
  keywordstyle=[3]\usefont{OT1}{cmtt}{m}{n},%\textit
  alsoletter={\#,\&},%
  keywords=[1]{not,from,import,def,if,else,return,while,break,and,or,for,in,del,and,class},%
  keywords=[2]{\#const,\#show,\#minimize,\#base,\#theory,\#count,\#external,\#program,\#script,\#end,\#heuristic,\#edge,\#project,\#show},%
  keywords=[3]{&,&dom,&sum,&diff,&show},%
  morecomment=[l]{\#\ },%
  morecomment=[l]{\%\ },%
  commentstyle={\color{darkgray}}%
}

\title[{\textit{Clingo} goes Linear Constraints over Reals and Integers}]{\textit{Clingo} goes Linear Constraints\\ over Reals and Integers}

\author[Tomi Janhunen et al.]{%
  Tomi Janhunen
  \\
  Aalto University
  \and 
  Roland Kaminski
  \\
  University of Potsdam
  \and
  Max Ostrowski
  \\
  University of Potsdam
  \and 
  Torsten Schaub
  \\
  INRIA Rennes and University of Potsdam
  \and
  Sebastian Schellhorn
  \\
  University of Potsdam
  \and
  Philipp Wanko
  \\
  University of Potsdam}

\submitted{[n/a]}
\revised{[n/a]}
\accepted{[n/a]}

\begin{document}

\maketitle

\begin{abstract}
The recent series~5 of the ASP system \clingo{} provides generic means to enhance basic Answer Set Programming (ASP)
with theory reasoning capabilities.
We instantiate this framework with different forms of linear constraints and elaborate upon its formal properties.
Given this, we discuss the respective implementations,
and present techniques for using these constraints in a reactive context.
%and conduct an experimental analysis with related systems.
More precisely,
we introduce extensions to \clingo{} with
difference and linear constraints over integers and reals, respectively, 
and realize them in complementary ways.
Finally, 
we empirically evaluate the resulting \clingo{} derivatives \clingod{dl} and \clingod{lp} on common language fragments and contrast them to related ASP systems.

\smallskip\noindent
{\em This paper is under consideration for acceptance in TPLP.}
\end{abstract}

%%% Local Variables: 
%%% mode: latex
%%% TeX-master: "paper"
%%% End: 

\section{Introduction}\label{sec:introduction}

Answer Set Programming (ASP;~\cite{baral02a}) has become an established paradigm for knowledge representation and reasoning,
in particular, when it comes to solving knowledge-intense combinatorial (optimization) problems.
Despite its versatility,
however, ASP falls short in handling non-Boolean constraints, such as linear constraints over unlimited integers or reals.
This shortcoming was broadly addressed in the recent \clingo~5 series~\cite{gekakaosscwa16a} by
providing generic means for incorporating theory reasoning.
They span from theory grammars for seamlessly extending \clingo's input language with theory expressions to 
a simple interface for integrating theory propagators into \clingo's solver component.

We instantiate this framework with different forms of linear constraints and elaborate upon its formal properties.
Given this, we discuss the respective implementations,
and present techniques for using these constraints in a reactive context.
In more detail,
we introduce extensions to \clingo{} with
difference and linear constraints over integers and reals, respectively, 
and realize them in complementary ways.
For handling difference constraints,
we provide customized implementations of well-established algorithms in \python{} and \cpp,
while we use \clingo's \python{} API to connect to off-the-shelf linear programming solvers, viz.\ \cplex{} and \lpsolve,
to deal with linear constraints.
In both settings, we support integer as well as real valued variables.
For a complement, we also consider \clingcon, a derivative of \clingo, 
integrating constraint propagators for handling linear constraints over integers at a low-level.
While this fine integration must be done at compile-time, the aforementioned \python{} extensions are added at run-time.
Our empirical analysis complements the study in~\cite{liesus16a} 
with experimental results on our new systems \clingod{dl} and \clingod{lp}.
Finally, we provide a comparison of different semantic options for integrating theories into ASP
and a systematic overview of the various features of state-of-the-art ASP systems handling linear constraints.

%%% Local Variables: 
%%% mode: latex
%%% TeX-master: "paper"
%%% End: 

\section{Answer Set Programming with Linear Constraints}\label{sec:background}

Our paper centers upon the theory reasoning capabilities of \clingo{} that allow us to extend ASP with linear constraints,
also referred to as \ASPm{lc}.
We focus below on the corresponding syntactic and semantic features, 
and refer the reader to the literature for an introduction to the basics of ASP. 

We consider (disjunctive) \emph{logic programs with linear constraints}, for short%
\footnote{We keep using the prefix `\emph{lc-}' throughout as a shorthand for concepts related to linear constraints.}
\emph{lc-programs},
over sets $\mathcal{A}$ and $\mathcal{L}$ of ground \emph{regular} and \emph{linear constraint atoms},
respectively.
An expression is said to be \emph{ground}, if it contains no ASP variables.
Accordingly, such programs consist of \emph{rules} of the form
\begin{lstlisting}[mathescape,numbers=none]
   a$_1$;...;a$_m$ :- a$_{m+1}$,...,a$_n$,not a$_{n+1}$,...,not a$_o$
\end{lstlisting}
where each \lstinline[mathescape]{a$_i$} is either 
a {regular atom} in $\mathcal{A}$ of form \lstinline[mathescape]{p(t$_1$,...,t$_k$)}
such that all \lstinline[mathescape]{t$_i$} are ground terms
or 
an {lc-atom} in $\mathcal{L}$ of form%
\footnote{In \clingo, theory atoms are preceded by `\texttt{\&}'.}
`\lstinline[mathescape]@&sum{a$_1$*x$_1$;$\dots$;a$_l$*x$_l$}<=k@' 
that stands for the linear constraint
\(
a_1\cdot x_1+\dots+a_l\cdot x_l\leq k
\).
All \lstinline[mathescape]{a$_i$} and \lstinline[mathescape]{k} are finite sequences of digits with at most one dot%
\footnote{In the input language of \clingo, sequences containing dots must be quoted to avoid clashes.} 
and represent real-valued coefficients $a_i$ and $k$.
Similarly all \lstinline[mathescape]{x$_i$} stand for the real (or integer) valued variables $x_i$.
As usual, \lstinline[mathescape]{not} denotes (default) \emph{negation}.
A rule is called a \emph{fact} if ${m,o}=1$,
\emph{normal} if $m=1$, and 
an \emph{integrity constraint} if $m=0$.
A linear constraint of form
\(
x_1-x_2\leq k
\)
is called a \emph{difference constraint},
and represented as
`\lstinline[mathescape]@&sum{x$_1$; -1*x$_2$}<=k@'
(or `\lstinline[mathescape]@&diff{x$_1$-x$_2$}<=k@' in pure difference logic settings). 

To ease the use of ASP in practice, 
several extensions have been developed. 
First of all, rules with ASP variables are viewed as shorthands for the set of their ground instances.
Further language constructs include
\emph{conditional literals} and \emph{cardinality constraints} \cite{siniso02a}.
The former are of the form
\lstinline[mathescape]{a:b$_1$,...,b$_m$},
the latter can be written as
\lstinline[mathescape]+s{d$_1$;...;d$_n$}t+,
where \lstinline{a} and \lstinline[mathescape]{b$_i$} are possibly default-negated (regular) literals  % for $0\leq i\leq m$,
and each \lstinline[mathescape]{d$_j$} is a conditional literal; % for $1\leq i\leq n$;
\lstinline{s} and \lstinline{t} provide optional lower and upper bounds on the number of satisfied literals in the cardinality constraint.
We refer to \lstinline[mathescape]{b$_1$,...,b$_m$} as a \emph{condition},
and call it \textit{static} if it is evaluated during grounding, otherwise it is called \textit{dynamic}.
The practical value of such constructs becomes apparent when used with ASP variables. 
For instance, a conditional literal like
\lstinline[mathescape]{a(X):b(X)}
in a rule's antecedent expands to the conjunction of all instances of \lstinline{a(X)} for which the corresponding instance of \lstinline{b(X)} holds.
Similarly,
\lstinline[mathescape]+2{a(X):b(X)}4+
is true whenever at least two and at most four instances of \lstinline{a(X)} (subject to \lstinline{b(X)}) are true.
%
% Finally, objective functions minimizing the sum of weights $w_i$ subject to condition $c_i$ are expressed as
% \lstinline[mathescape]!#minimize{$w_1$:$c_1$;$\dots$;$w_n$:$c_n$}!.
% \lstinline[mathescape]!#minimize{$w_1$@$l_1$:$c_1$;$\dots$;$w_n$@$l_n$:$c_n$}!.
% Lexicographically ordered objective functions are (optionally) distinguished via levels indicated by $l_i$.

Likewise,
\clingo's syntax of linear constraints offers several convenience features.
As above,
elements in linear constraint atoms can be conditioned (and use ASP variables),
viz.\
`\lstinline[mathescape]@&sum{a$_1$*x$_1$:c$_1$;...;a$_l$*x$_l$:c$_n$}<=k@'
where each \lstinline[mathescape]{c$_i$} is a condition.
As above, the usage of ASP variables allows for forming arbitrarily long expressions
(cf.\ Listing~\ref{encoding:yale}).
That is, by using static or dynamic conditions,
we may formulate linear constraints that are determined relative to a problem instance during grounding
and even dynamically during solving, respectively. 
Also, linear constraints can be formed with further relations, viz.\
\texttt{>=},
\texttt{<},
\texttt{>},
\texttt{=},
and
\texttt{!=}.
Moreover, the theory language for linear constraints offers a domain declaration for real variables,
`\lstinline[mathescape]@&dom{lb..ub}=x@'
expressing that all values of \texttt{x} must lie between \texttt{lb} and \texttt{ub}, inclusive.
And finally the maximization (or minimization) of an objective function can be expressed with
\lstinline[mathescape]@&maximize{a$_1$*x$_1$:c$_1$;...;a$_l$*x$_l$:c$_n$}@
(or by \texttt{minimize}).
The full theory grammar for linear constraints over reals is available at~\url{https://potassco.org/clingo/examples}.

Semantically, a logic program induces a set of \emph{stable models},
being distinguished models of the program determined by the stable models semantics~\cite{gellif91a}.
To extend this concept to logic programs with linear constraints,
we follow the approach of lazy theory solving~\cite{baseseti09a}.
We abstract from the specific semantics of a theory by considering the lc-atoms representing the underlying linear constraints.
The idea is that a regular stable model $X$ of a program over $\mathcal{A}\cup\mathcal{L}$ is only valid wrt the theory,
if the constraints induced by the truth assignment to the lc-atoms in $\mathcal{L}$ are satisfiable in the theory.
In our setting, 
this amounts to finding an assignment of reals (or integers) to all numeric variables that 
satisfies a set of linear constraints induced by $X\cap\mathcal{L}$.
Although this can be done in several ways, as detailed below,
let us illustrate this by a simple example.
The (non-ground) logic program containing the fact 
`\lstinline[mathescape]{a("1.5").}'
along with the rule
`\lstinline[mathescape]@&sum{R*x}<=7 :- a(R).@'
has the regular stable model \lstinline[mathescape]@$\{$a("1.5")$,\;$&sum{"1.5"*x}<=7$\}$@.
Here, we easily find an assignment, e.g.\ $\{x\mapsto 4.2\}$,
that satisfies the only associated linear constraint `$1.5*x\leq 7$'.

In what follows, we make this precise by instantiating 
the general framework of logic programs with theories in~\cite{gekakaosscwa16a}
to the case of linear constraints over reals and integers (and so difference constraints).
Also, we focus on one theory at a time.
Thereby,
our emphasis lies on the elaboration of alternative semantic options for stable models with linear constraints,
which pave the way for different implementation techniques discussed in Section~\ref{sec:system}.

We use the following notation.
Given a rule $r$ as above,
we call $\{\mathtt{a}_1,\dots,\mathtt{a}_m\}$ % and $\{\mathtt{a}_{m+1},\dots,\mathtt{a}_n,\mathtt{not}\,\mathtt{a}_{n+1},\dots,\mathtt{not}\,\mathtt{a}_o\}$
its \emph{head} % and \emph{body} 
and denote it % them
by \head{r}. % and \body{r}.
Furthermore, we define
\(
\head{P}=\bigcup_{r\in P}\head{r}
\).
% we let \head{P} stand for all head atoms in program $P$.

First of all,
we may distinguish whether linear constraints are only determined outside or additionally inside a program.
Accordingly,
we partition $\mathcal{L}$ into
\emph{defined} and \emph{external} lc-atoms,
namely $\mathcal{L}\cap\head{P}$ and $\mathcal{L}\setminus\head{P}$, 
respectively.\footnote{This distinction is analogous to that between head and input atoms,
  defined via rules or \lstinline{#external} directives \cite{gekakasc14b}, respectively.}
While external lc-atoms must only be satisfied by the respective theory,
defined ones must additionally be derivable through rules in the program.
The second distinction is about the logical correspondence between theory atoms and theory constraints.
To this end,
we partition $\mathcal{L}$ into
\emph{strict} and \emph{non-strict} lc-atoms,
denoted by $\mathcal{L}^\leftrightarrow$ and $\mathcal{L}^\rightarrow$, respectively.
The strict correspondence requires
a linear constraint to be satisfied 
\textit{iff}
the associated lc-atom in $\mathcal{L}^\leftrightarrow$ is true.
A weaker condition is imposed in the non-strict case.
Here, a linear constraint must hold 
\textit{only if}
the associated lc-atom in $\mathcal{L}^\rightarrow$ is true.
Thus, only lc-atoms in $\mathcal{L}^\rightarrow$ assigned true impose requirements, 
while constraints associated with falsified lc-atoms in $\mathcal{L}^\rightarrow$ are free to hold or not.
However, by contraposition, a violated constraint leads to a false lc-atom.

Different combinations of such correspondences are possible,
and we may even treat some constraints differently than others.
In view of this, we next provide an extended definition of stable models that accommodates all above correspondences.
Following~\cite{gekakaosscwa16a},
we accomplish this by mapping the semantics of lc-programs back to regular stable models.
To this end,
we abstract from the actual constraints and identify a solution with a set of linear constraint atoms.
More precisely,
we call
\(
\mathcal{S}\subseteq\mathcal{L}
\)
a \emph{linear constraint solution},
if there is an assignment of reals (or integers) to all real (integer) valued variables represented in $\mathcal{L}$ that 
\begin{enumerate}
 \item[(i)]  \label{en:lcsol1} satisfies all linear constraints associated with strict and non-strict lc-atoms in $\mathcal{S}$ and
 \item[(ii)] \label{en:lcsol2} falsifies all linear constraints associated with strict lc-atoms in $\mathcal{L}^\leftrightarrow\setminus\mathcal{S}$.
\end{enumerate}

Then, we define a set $X\subseteq\mathcal{A}\cup\mathcal{L}$
as an \emph{lc-stable model} of an lc-program~$P$,
if
there is some lc-solution $\mathcal{S}\subseteq\mathcal{L}$ such that
$X$ is a (regular) stable model of the logic program
\newcommand{\code}[1]{\text{\texttt{#1}}}
\begin{align}
\hspace{-10pt}
P
&{}\cup
\{\code{a.}       \mid \code{a}\in (\mathcal{L}^\leftrightarrow\setminus\head{P})\cap \mathcal{S}\}
\cup
\{\code{:- not a.}\mid \code{a}\in (\mathcal{L}^\leftrightarrow\cap\head{P})\cap \mathcal{S}\}
\label{eq:stable:strict}
\\
 &{}\cup
\{\code{\{a\}.}   \mid \code{a}\in (\mathcal{L}^\rightarrow\setminus\head{P})\cap \mathcal{S}\}
\cup
\{\code{:- a.}    \mid \code{a}\in (\mathcal{L}\cap\head{P})\setminus \mathcal{S}\}\rlap{.}
\label{eq:stable:non-strict}
\end{align}
The rules added to~$P$ % in~\eqref{eq:stable:strict} and~\eqref{eq:stable:non-strict}
express conditions aligning the lc-atoms in $X\cap\mathcal{L}$ with a corresponding lc-solution~$\mathcal{S}$.
To begin with, 
the set of facts on the left in~\eqref{eq:stable:strict} makes sure that 
all lc-atoms in~$\mathcal{S}$ that are external and strict % , viz.\ in $\mathcal{L}^\leftrightarrow\setminus\head{P}$,
also belong to~$X$.
Unlike this, the corresponding set of choice rules in~\eqref{eq:stable:non-strict}
merely says that non-strict external lc-atoms from~$\mathcal{S}$ may be included in~$X$ or not.
The integrity constraints in~\eqref{eq:stable:strict} and~\eqref{eq:stable:non-strict}
take care of defined lc-atoms, viz.\ the ones in~$\head{P}$.
The set in~\eqref{eq:stable:strict} again focuses on strict lc-atoms and
stipulates that the ones from~$\mathcal{S}$ are included in~$X$ as well.
Finally, for both strict and non-strict defined lc-atoms,
the integrity constraints in~\eqref{eq:stable:non-strict} assert the falsity
of atoms that are not in~$\mathcal{S}$.

%%% Local Variables: 
%%% mode: latex
%%% TeX-master: "paper"
%%% End: 

In what follows,
we elaborate upon the formal relationships among the different types of lc-atoms.
To this end,
we distinguish four homogeneous settings, in which all lc-atoms are either
\lcatype{defined}{strict},
\lcatype{defined}{non-strict},
\lcatype{external}{strict}, or
\lcatype{external}{non-strict}, respectively.
We use the following notation.
For an lc-program $P$ over $\mathcal{A}\cup\mathcal{L}$ and an lc-solution $\mathcal{S}\subseteq\mathcal{L}$,
we define \PS{P}{S} as the extension of program $P$ given in (\ref{eq:stable:strict}) and (\ref{eq:stable:non-strict}).
Also,
we let \smset{P} denote the set of (regular) stable models of program $P$ over $\mathcal{A}\cup\mathcal{L}$,
and
\(
\smsetlc{P}=
\bigcup_{\mathcal{S}\subseteq\mathcal{L}\text{ lc-solution}}
\smset{\PS{P}{S}}
\)
its set of lc-stable models.
Note that the respective semantic setting is determined by the type of lc-atoms in $\mathcal{L}$.
In fact, two syntactically equivalent lc-programs may yield different lc-models in different settings.
This is made precise in the following proposition.
% --------------------------------------------------------------------------------
\begin{theorem}\label{thm:settings}
  Let $P$ be an lc-program over $\mathcal{A}\cup\mathcal{L}$
  and $P'$   an lc-program over $\mathcal{A}\cup\mathcal{L}'$
  such that $P=P'$.
  \begin{enumerate}
  \item \label{th:dsr} % X_d subseteq X_r
    If
    \(
    \mathcal{L}=\mathcal{L}\cap\head{P}
    \),
    then
    \(
    \smsetlc{P}\subseteq\smset{P}
    \)
  \item \label{th:rsen} % X_r subseteq X_en
    If
    \(
    \mathcal{L}=\mathcal{L}^\rightarrow\setminus\head{P}
    \),
    then
    \(
    \smset{P}\subseteq\smsetlc{P}
    \)
  \item \label{th:ssn} % X_s subseteq X_n
    If
    \(
    \mathcal{L}'=\mathcal{L}'^\rightarrow
    \),
    then
    \(
    \smsetlc{P}\subseteq\smsetlc{P'}
    \)
  \end{enumerate}
\end{theorem}
% --------------------------------------------------------------------------------
%
Note that $P=P'$ also makes $\mathcal{L}$ and $\mathcal{L}'$ syntactically equivalent,
although they may represent different types of lc-atoms.
The above results draw on the observation that if all atoms in $\mathcal{L}'$ are non-strict, then
\(
\{\mathcal{S}\subseteq\mathcal{L} \mid \mathcal{S}\text{ is an lc-solution}\}
\subseteq
\{\mathcal{S}\subseteq\mathcal{L}'^\rightarrow \mid \mathcal{S}\text{ is an lc-solution}\}
\).
This is because the former set of lc-solutions need to satisfy at least condition~(i)
while the latter must only satisfy~(i).
Note that Proposition~\ref{thm:settings} does not just apply to \ASPm{lc} but to ASP modulo arbitrary theories.

In more detail,
Proposition~\ref{thm:settings}.\ref{th:dsr} expresses that each      lc-stable model is also a  regular stable model
in a setting involving defined lc-atoms only.
Conversely,
Proposition~\ref{thm:settings}.\ref{th:rsen} expresses that each regular stable model is also an      lc-stable model
in the \lcatype{external}{non-strict} setting.
Proposition~\ref{thm:settings}.\ref{th:ssn} portrays that handling lc-atoms in a strict or non-strict way
may lead to fewer (or equal) lc-stable models than treating them just in a non-strict way.

In contrast to the observations of Proposition~\ref{thm:settings},
the following proposition tells us that regular and lc-stable models are in general incomparable in the \lcatype{external}{strict} setting.
\begin{theorem}\label{thm:settings2} % X_es not subseteq X_r   and   X_r not subseteq X_es
    There exist lc-programs $P$ over $\mathcal{A}\cup\mathcal{L}$ with
    \(
    \mathcal{L}=\mathcal{L}^\leftrightarrow\setminus\head{P}
    \),
    so that
    \(
    \smset{P}\not\subseteq\smsetlc{P}
    \)
    or
    \(
    \smsetlc{P}\not\subseteq\smset{P}
    \).
\end{theorem}
% --------------------------------------------------------------------------------
This results from the fact that
the treatment of strict lc-atoms may prune regular stable models
and, on the other hand, the pure external evaluation of lc-atoms may induce additional stable models.

Now that we have explored the formal correspondence among the alternative settings,
let us discuss their appropriateness for \ASPm{lc}.
To this end, let us consider two examples.

We first asses the two defined settings. Modifying our above example, let $P_1$ be
\begin{lstlisting}[numbers=none,mathescape]
{a("1.5")}.  &sum{"1.5"*x}<=7 :- a("1.5").  &sum{x}<"4.5".
\end{lstlisting}
This logic program has two regular stable models
$X_1=\{$\lstinline@ &sum{x}<"4.5" @$\}$ and
$X_2=\{$\lstinline@ a("1.5"),@ \lstinline@&sum{"1.5"*x}<=7, &sum{x}<"4.5" @$\}$.

Let us first consider the \lcatype{defined}{strict} case,
in which the lc-atoms \lstinline@&sum{"1.5"*x}<=7@ and \lstinline@&sum{x}<"4.5"@
belong to $\mathcal{L}^\leftrightarrow\cap\head{P}$.
Then, $\mathcal{S}_a=\emptyset$ is an lc-solution, since both $1.5*x\leq7$ and $x<4.5$ can be falsified.
However,
the resulting program $\PS{P_1}{\mathcal{S}_\mathit{a}}$ contains rules
`\lstinline[mathescape]@&sum{x}<"4.5".@' and `\lstinline[mathescape]@:- &sum{x}<"4.5".@'
and thus admits no regular stable model.
The same result is obtained for
$\mathcal{S}_b=\{$\lstinline@ &sum{"1.5"*x}<=7 @$\}$.
Unlike this,
$\mathcal{S}_c=\{$\lstinline@ &sum{x}<"4.5" @$\}$
is no lc-solution
although it appears to support $X_1$ as an lc-model.
In a strict setting, an \textit{iff} correspondence is imposed between lc-atoms and their associated linear constraints.
This excludes $\mathcal{S}_c$ as an lc-solution,
since there is no real-valued assignment satisfying $x<4.5$ while falsifying $1.5*x\leq7$.
This situation is caused by the non-derivability of lc-atom \lstinline@&sum{"1.5"*x}<=7@,
which is in turn falsified by the stable models semantics.
The strict interpretation of the lc-atom then requires the falsification of $1.5*x\leq7$.
Finally, $\mathcal{S}_d=\{$\lstinline@ &sum{x}<"4.5", &sum{"1.5"*x}<=7 @$\}$ is another lc-solution.
Given that
\(
\PS{P_1}{\mathcal{S}_\mathit{d}}=P_1\;\cup\;$\lstinline[mathescape]@$\{$ :- not &sum{"1.5"*x}<=7.  :- not &sum{x}<"4.5". @$\}
\)
has the regular stable model $X_2$, we establish that $X_2$ is the only lc-stable model of $P_1$.

This example illustrates that strict lc-atoms impose a rather strong connection to their associated constraints in a defined setting.
Hence, let us consider next the above example in a \lcatype{defined}{non-strict} setting,
requiring merely an \textit{only if} condition between constraints and their lc-atoms.
Now, $\mathcal{S}_c=\{$\lstinline@ &sum{x}<"4.5" @$\}$ is an lc-solution since $1.5*x\leq7$
must not be falsified.
Accordingly, the regular stable model $X_1$ of
\(
\PS{P_1}{\mathcal{S}_\mathit{c}}=P_1\;\cup\;$\lstinline[mathescape]@$\{$ :- &sum{"1.5"*x}<=7. @$\}
\)
attests that $X_1$ is also an lc-stable model of $P_1$.
The other lc-solutions yield the same results as above.

Next, let us analyze the two external settings.
For this,
let the lc-program $P_2$ be
\begin{lstlisting}[numbers=none,mathescape]
:- not &sum{x}<"4.5".  a("1.5") :- &sum{"1.5"*x}<=7.
\end{lstlisting}
admitting no regular stable models, due to the included integrity constraint.

First, we examine the \lcatype{external}{non-strict} setting.
In this case,
each combination of the lc-atoms \lstinline@&sum{"1.5"*x}<=7@ and \lstinline@&sum{x}<"4.5"@
in $\mathcal{L}^\rightarrow\setminus\head{P}$ results in an lc-solution.
However,
the existence of lc-stable models depends upon the presence of lc-atom \lstinline@&sum{x}<"4.5"@.
Lc-models are obtained if it is included, otherwise the integrity constraint in $P_2$ denies them.
The lc-solution $\mathcal{S}_a=\{$\lstinline@ &sum{x}<"4.5" @$\}$ results in the identical lc-stable model.
Note that all underlying assignments must satisfy $x<4.5$ and hence $1.5*x\leq7$.
However, the non-strict nature of \lstinline@&sum{"1.5"*x}<=7@ leaves its truth value open.
Thus, stable model semantics may set it to false and \lstinline@a("1.5")@ is not obtained
although the actual constraint $1.5*x\leq7$ in the rule body in $P_2$ is satisfied.
Similarly,
the lc-solution $\mathcal{S}_b=\{$\lstinline@ &sum{x}<"4.5", &sum{"1.5"*x}<=7 @$\}$
induces the same counter-intuitive lc-model $\{$\lstinline@ &sum{x}<"4.5" @$\}$
along with a second, arguably more intuitive lc-model
\(
\{$\lstinline[mathescape]@ a("1.5"), &sum{"1.5"*x}<=7, &sum{x}<"4.5" @$\}
\).

The previous discussion has revealed that non-strict lc-atoms may ignore information induced by the theory in an external setting.
This lack is compensated in an \lcatype{external}{strict} setting by the above condition (ii)
and the resulting assertion of lc-atoms representing satisfied constraints in (\ref{eq:stable:strict}).
Accordingly,
\(
\{$\lstinline[mathescape]@ a("1.5"), &sum{"1.5"*x}<=7, &sum{x}<"4.5" @$\}
\)
is the only lc-stable model of $P_2$.
By interpreting both lc-atoms in a strict manner,
the inclusion of \lstinline[mathescape]@&sum{x}<"4.5"@ entails that of
\lstinline@&sum{"1.5"*x}<=7@ as well.
Hence, the singleton $\{$\lstinline@ &sum{x}<"4.5" @$\}$ cannot be an lc-model of $P_2$
in a \lcatype{external}{strict} setting.

The previous discussion has shown that certain semantic combinations are more appropriate for treating linear constraints than others.
This may be different for other theories.
We have seen that a \lcatype{defined}{strict} interpretation of lc-atoms may be overly strong,
since the non-derivability of lc-atoms may severely restrict real-valued assignments.
Conversely, the \lcatype{external}{non-strict} treatment of lc-atoms may be too weak,
since it admits real-valued variable assignments satisfying constraints that are not
reflected in the corresponding lc-stable models.
As a consequence, we focus in what follows on the \lcatype{external}{strict} and \lcatype{defined}{non-strict} settings for lc-atoms.

Finally, let us comment on the usability of both types of lc-atoms.
Their \lcatype{external}{strict} interpretation allows for deriving information from the respective theory.
This generates some overhead since the corresponding propagators have to deal with
two relations between lc-atoms and their associated constraints.
This approach is advantageous in our planning example in Section~\ref{sec:multishot}, 
where \lcatype{external}{strict} lc-atoms allow us 
to naturally express goal conditions as integrity constraints.
Conversely, we face the following difficulties.
First, defined lc-atoms must also occur in some rule head, which is rarely the case with goal conditions.
Second, non-strict lc-atoms may be false although the actual constraint is satisfied.
On the other hand, 
in the \lcatype{defined}{non-strict} setting,
the stable model semantics delineates the effective set of constraints that needs to be satisfied.
False lc-atoms are considered as unknown and can therefore be disregarded by the corresponding propagators.
We draw on this in our scheduling encodings where it halves the number of constraints and helps with faster propagation via the program's completion.
The impact of this is investigated in Section~\ref{sec:experiments}.
As a rule of thumb,
the choice between both settings depends on who should be in charge of delineating the set of constraints in focus.
If this is the theory propagator, an \lcatype{external}{strict} setting is preferable,
since the strict correspondence induces the relevant lc-atoms without any interference with derivable lc-atoms.
If this is the actual ASP system, a \lcatype{defined}{non-strict} setting is favorable,
in which derivable lc-atoms delineate the set of constraints checked by the constraint propagator.
%
%%% Local Variables:
%%% mode: latex
%%% TeX-master: "paper"
%%% End:

\section{Multi-Shot ASP Solving with Linear Constraints}
\label{sec:multishot}
Multi-shot solving~\cite{gekakasc14b} is about solving continuously changing logic programs in an operative way.
This can be controlled via reactive procedures that loop on solving while reacting, for instance, to outside changes or previous solving results.
These reactions may entail the addition or retraction of rules that the operative approach can accommodate by leaving the unaffected program parts
intact within the solver.
This avoids re-grounding and benefits from heuristic scores and nogoods learned over time.
In fact, 
evolving logic programs with linear constraints can be extremely useful in dynamic applications, for example, to %:
add new resources in a planning domain,
or to set the value of an observed variable measured using sensors.
The abstraction from actual constraints to constraint atoms
allows us to easily extend multi-shot solving to lc-programs.

To illustrate how seamlessly our systems \clingod{dl} and \clingod{lp} support multi-shot solving,
we apply the exemplary \python\ script, shipped with \clingo\ to illustrate incremental solving,
to model the spoiling Yale shooting scenario~\cite{caotpo00a}.
Multi-shot solving in \clingo\ relies on two directives (cf.~\cite{gekakasc14b}),
the \texttt{\#program} directive for regrouping rules
and
the \texttt{\#external} directive for declaring atoms
as being external to the program at hand.
The truth value of such external atoms can be set via \clingo's API\@.
The aforementioned \python\ script loops over increasing integers until a stop criterion is met.
It presupposes three groups of rules declared via \texttt{\#program} directives.
At step 0 the programs named \texttt{base} and \texttt{check(n)} are grounded and solved for $\texttt{n}=0$.
Then, in turn programs \texttt{check(n)} and \texttt{step(n)} are added for $\texttt{n}>0$, grounded, and the resulting overall program solved.
% Other names and components are definable by appropriate changes to the script.
% Stop criteria can be the satisfiability or unsatisfiability of the respective program at each iteration.
In addition, at each step $\texttt{n}$ an external atom \texttt{query(n)} is introduced;
it is set to true for the current iteration $\texttt{n}$ and false for all previous instances with smaller integers than $\texttt{n}$.
We refer the reader to~\cite{gekakasc14b} for further details on the \python{} part.
Notably, for dealing with lc-programs,
we can use the exemplary \python\ script as is---once the respective propagator is registered with the solver.

In the spoiled Yale shooting scenario~\cite{caotpo00a},
we have a gun and two actions, viz.\ load and shoot.
If we load, the gun becomes loaded.
If we shoot, it kills the turkey, if the gun was loaded for no more than 35 minutes.
Otherwise, the gun powder is spoiled.
We model this planning problem in \ASPm{lc}.
% --------------------------------------------------------------------------------
\begin{lstlisting}[caption={Spoiled Yale shooting instance},float=ht,label=encoding:yalebase,basicstyle=\ttfamily\footnotesize]
#include "incmode_lc.lp".

#program base.
  action(load).       action(shoot).      action(wait).
duration(load,25).  duration(shoot,5).  duration(wait,36).
unloaded(0).
&sum { at(0) } = 0.
&sum { armed(0) } = 0.
\end{lstlisting}
% --------------------------------------------------------------------------------
We start by including the incremental \python\ program,
the grammar, and the propagator for linear constraints in the first line of Listing~\ref{encoding:yalebase}.%
\footnote{For uniformity, we use semi-colons '\texttt{;}' rather than '\texttt{,}' for separating body elements.}
This listing is the base program.
All actions and their durations are introduced in Lines~4 and~5.
At the initial situation, the gun is unloaded (Line~6).
Line~7 and~8 initialize integer variables \texttt{at(0)} and \texttt{armed(0)} with 0 (see below).
% --------------------------------------------------------------------------------
\begin{lstlisting}[caption={Spoiled Yale shooting scenario},float=ht,label=encoding:yale,basicstyle=\ttfamily\footnotesize]
#program step(n).
1 { do(X,n) : action(X) } 1.
&sum { at(n); -1*at(N') } = D :- do(X,n); duration(X,D); N'=n-1.

loaded(n)   :- loaded(n-1); not unloaded(n).
unloaded(n) :- unloaded(n-1); not loaded(n).
dead(n)     :- dead(n-1).

&sum { armed(n) } = 0 :- unloaded(n-1).
&sum { armed(n); -1*armed(N') } = D :- do(X,n); duration(X,D); N'=n-1; loaded(N').

loaded(n)   :- do(load,n).
unloaded(n) :- do(shoot,n).
dead(n)     :- do(shoot,n); &sum { armed(n) } <= 35.
:- do(shoot,n); unloaded(n-1).
\end{lstlisting}
% --------------------------------------------------------------------------------
Listing~\ref{encoding:yale} gives the dynamic part of the problem;
it is grounded for each step \texttt{n}.
Line~2 enforces that exactly one action is done per step.
The exact times at which each step takes place is captured by the integer variables \texttt{at(n)}.
The difference between two consecutive time steps is the duration
of the respective action (Line~3).
The next three lines make the fluents inertial, viz.\
the gun stays loaded/unloaded if it was loaded/unloaded before,
and the turkey remains dead.
Lines~9 and~10 use the integer variable \texttt{armed(n)}
to describe for how long the weapon has been loaded at step \texttt{n}.
Whenever it is unloaded, \texttt{armed(n)} is 0,
otherwise it is increased by the duration of the last action.
The following four lines (12--15) encode the conditions and effects of the actions.
When we load the gun, it becomes loaded; when we shoot, it becomes unloaded.
If we shoot and the gun was loaded for no longer than 35 minutes (and thus the gun powder is unspoiled),
the turkey is dead.
The last line ensures that we cannot shoot if the gun is not loaded.
Together with the initial situation and the actions from Listing~\ref{encoding:yalebase}
this encodes the spoiled Yale shooting problem,
and any solution represents an executable plan.
% --------------------------------------------------------------------------------
\begin{lstlisting}[caption={Query for the spoiled Yale Shooting Scenario.},float=ht,label=encoding:yalequery,basicstyle=\ttfamily\footnotesize]
#program check(n).
:- not dead(n); query(n).
:- not &sum { at(n) } <= 100; query(n).
:- do(shoot,n); not &sum { at(n) } > 35.
\end{lstlisting}
% --------------------------------------------------------------------------------
Listing~\ref{encoding:yalequery} adds a query to our problem.
In Line~2 we require that the turkey is dead at step \texttt{n}.
As this constraint is subject to the external atom \texttt{query(n)},
it is only active at solving step \texttt{n}.
The next line ensures that we kill the turkey within 100 minutes.
And as an additional constraint,
we added some preparation time such that we are not allowed 
to shoot in the first 35 minutes.
It is possible to solve this problem within three steps.
There exist two solutions at this time point,
one of them containing
\texttt{unloaded(0)}, \texttt{do(wait,1)}, \texttt{unloaded(1)},
\texttt{do(load,2)}, \texttt{loaded(2)},
\texttt{do(shoot,3)}, \texttt{unloaded(3)}, \texttt{dead(3)}.
That is, we simply wait before loading and shooting.
The second solution loads the gun instead of waiting,
thus loading the gun twice before shooting.

%%% Local Variables:
%%% mode: latex
%%% TeX-master: "paper"
%%% End:

\section{\clingo{} derivatives and related systems}\label{sec:system}

In this section, we give an overview of systems extending ASP with linear constraints.
We start with our own systems \clingod{dl} and \clingod{lp}
both relying upon \clingo's interface for theory propagators.
We also include \clingcon, since it is based on a much lower level API
using the internal functions of \clingo{} (and \clasp) in \cpp.
While \clingcon{} implements a highly sophisticated system using advanced preprocessing and solving techniques,
the \python{} variants of \clingod{dl} and \clingod{lp} provide easily modifiable and maintainable propagators for difference and linear constraints,
respectively.
This carries over to the \cpp{} variant of \clingod{dl} since the \cpp{} and \python{} API share the same functionality. 
% --------------------------------------------------------------------------------
\begin{table}%
\caption{Feature comparison}%
\label{tab:features}%
\center%
\begin{threeparttable}
\begin{tabular}{@{}l@{}@{}c@{}@{}c@{}@{\;}c@{}@{}c@{}@{}c@{}@{}c@{}@{}c@{}@{}c@{}@{}c@{}}
\toprule
             & \python{} &   \cpp{} &            strict & non-strict & external &           defined &    n-ary &             reals &      optimization\\
\midrule
\clingod{dl} &  \cmark{} & \cmark{} & \cmark{}\tnote{1} &   \cmark{} & \cmark{} & \cmark{}          & \xmark{} & \cmark{}\tnote{2} & \cmark{}\tnote{3}\\
\clingod{lp} &  \cmark{} & \xmark{} & \cmark{}          &   \cmark{} & \cmark{} & \cmark{}          & \cmark{} & \cmark{}          & \cmark{}\tnote{4}\\
\clingcon{}  &  \xmark{} & \cmark{} & \cmark{}          &   \xmark{} & \cmark{} & \xmark{}\tnote{5} & \cmark{} & \xmark{}          & \cmark{}         \\
%\bottomrule
\end{tabular}%
\begin{tablenotes}\footnotesize
  \begin{minipage}{0.375\textwidth}
  \item[1] Only with \python{} API
  \item[2] Only for non-strict lc-atoms
  \item[3] Needs an additional plugin
  \end{minipage}%
  \begin{minipage}{0.6\textwidth}
  \item[4] Optimization is relative to stable models
  \item[5] Theory atoms in rule heads are shifted into negative body
  \item[] \
  \end{minipage}
\end{tablenotes}%
\end{threeparttable}%
\end{table}%
%
%%% Local Variables:
%%% mode: latex
%%% TeX-master: "../paper"
%%% End:

% --------------------------------------------------------------------------------
Table~\ref{tab:features} shows a comparative list of features for these systems.
The two flexible \clingo{} derivatives support all four combinations
of strict/non-strict and defined/external lc-atom types,
whereas \clingcon{} has a fixed one.
Also the bandwidth of supported constraints is different.
While \clingod{dl} only supports difference constraints,
the other two support n-ary linear constraints.
Notably, \clingod{dl} and \clingod{lp} support (approximations of) real numbers (see below).
Moreover, all three \clingo{} derivatives allow for optimizing objective functions over
numeric variables (in addition to optimization in ASP).
% All systems are freely available at \url{https://potassco.org/{clingoDL,clingoLP,clingcon}}.

\textbf{\clingod{dl}}
extends \clingo{} with difference constraints of the form $x-y\leq k$,
where $x$ and $y$ are integer (or real) variables and $k$ is an integer (real) constant.
Despite the restriction to two variables, 
they allow for naturally encoding timing related problems, as e.g., in scheduling, % or timetabling,
and are solvable in polynomial time.
%Syntactically, a difference constraint $x-y\leq k$ is represented by a difference constraint atom of the form `\lstinline[mathescape]@&diff{x-y}<=k@'.
%
\clingod{dl} uses \clingo{}'s theory interface to realize a stateful propagator
that checks during search whether the current set of implied difference constraints
is satisfiable~\cite{cotmal06a}.
To this end,
it makes use of the stateful nature of the theory interface
that allows 
for incrementally updating internal states
and thus for backtracking to previous states without having to rebuild the internal representation.
% \clingod{dl} is available using either the \python\ or \cpp\ interface at \url{https://github.com/potassco/clingoDL}.
%
By default, all difference constraint atoms are considered to be non-strict.
In this case, it is only necessary to keep track of lc-atoms that are assigned true
since only then the constraint is required to hold.
In the strict case,
false assignments to difference constraint atoms are considered as well.
This is done by adding $y-x\leq-k-1$ 
whenever `\lstinline[mathescape]@&diff{x-y}<=k@' is assigned false.
As a side-product of the satisfiability check,
an integer (real) assignment for all variables is obtained
and ultimately printed for all lc-stable models.
Usually, several or even an infinite number of assignments exist. % if the set of difference constraints is satisfiable.
The returned assignment is the one with the lowest sum of the absolute values of all variables.
For instance, in terms of scheduling problems, this amounts to scheduling each job as soon as possible.
%%%Besides binary constraints, it is possible to encode unary constraints
%%%by using the constant `\lstinline[mathescape]@0@', 
%%%where `\lstinline[mathescape]@0@' is treated as a dedicated variable being always $0$, 
%%%eg $x\leq10$ is encoded as `\lstinline[mathescape]@&diff{x-0} <= 10@'.\comment{maybe remove mathescape or use it in the constraint?}

\textbf{\clingod{lp}}
fully covers the extension of \ASPm{lc} described in Section~\ref{sec:background}.
This \clingo{} derivative accepts lc-atoms containing integer and real variables possibly subject to dynamic conditions.
That is, \clingod{lp} extends ASP with constraints as dealt with in Linear Programming (LP;~\cite{dantzig63a})
as well as according objective functions for optimization. 
In \clingod{lp}, the latter are subject to dynamic conditions and thus depend on the respective Boolean assignment
(as in regular ASP optimization).
As above,
the theory interface of \clingo{} is used to integrate a stateful propagator that checks during search 
the satisfiability of the current set of linear constraints.
Here, however, this is done with a generic interface to dedicated LP solvers,
currently supporting \cplex\ and \lpsolve.
(Note that both LP solvers do an exponential consistency check.) % use the simplex algorithm. \comment{P: ref to simplex maybe} 
The \python\ interfaces of \cplex\ and \lpsolve\ natively support relations $=$, $\geq$, and $\leq$. 
We add support for $<$, $>$, and $\neq$.
To this end, %similar to \cplex\ linearization, %\footnote{See~\url{https://www.ibm.com/support/knowledgecenter/SS9UKU\_12.4.0/com.ibm.cplex.zos.help/Parameters/topics/EpLin.html} for detailed information.} 
we translate $<$ and $>$ into $\leq$ and $\geq$ by subtracting or adding an $\varepsilon$ to the right-hand-side of a linear constraint, respectively.%
\footnote{This $\varepsilon$ can be configured using the command line and defaults to $10^{-3}$ (as in \cplex).}
Furthermore, $\neq$ is treated as a disjunction of $<$ and $>$.
By default, \clingod{lp} treats lc-atoms in a non-strict manner.
Thus only linear constraints represented by true lc-atoms are considered.
When treating them strictly, false lc-atoms are handled using the complementary relation.
In this case, 
the corresponding linear constraint is derived by using the complementary relation.
Notably, \clingod{lp} offers dynamic conditions in lc-atoms.
This allows for linear constraints of variable length even during search.
All conditions have to be decided before such a constraint is included in the consistency check.
Furthermore, \clingod{lp} updates its internal state incrementally 
but rebuilds the linear constraint system after backtracking to avoid accumulating rounding errors. 
%
%If the LP solver establishes inconsistency, a nogood is added  only consisting of literals associated with current linear constraints.
%Naively, all assigned lc-atoms and conditions may be gathered as reason for the inconsistency.
%But often only a subset of linear constraints is responsible for a conflict.  
%To this end, we identify a core conflict using the Irreducible Inconsistent Set (IIS) algorithm \cite{ostsch12a,loon81}.
Also, it uses an Irreducible Inconsistent Set algorithm~\cite{loon81}
for extracting minimal sets of conflicting constraints to support conflict learning in the ASP solver.
On the one hand, this extraction is expensive, on the other hand, such core conflicts may significantly reduce the search space. 
To control this trade-off, \clingod{lp} only enables this feature after a certain percentage of lc-atoms and conditions is assigned (by default 20\%).
Similarly,
frequent theory consistency checks are expensive and a conflict is less likely to be found within a small assignment;
accordingly, an analogous percentage based threshold allows for controlling their invocation (default 0\%). 
%
% \clingod{lp} is available at \url{https://github.com/potassco/clingoLP}.

\textbf{\clingcon} series~3 offers a \clingo-based ASP system with handcrafted propagators for constraints over integers~\cite{bakaossc16a};
it is implemented in \cpp{} and features a strict, external semantics.
Sophisticated preprocessing techniques are supported
and non-linear constraints such as the global \emph{distinct} constraint are translated into linear ones.
Integer variables are represented using the order encoding~\cite{crabak94a},
and customized propagators using state-of-the-art lazy nogood and variable generation are employed.
The propagators do not only ensure bound consistency on the variables
but also derive new bounds.
%This means that new constraints are derived.
Furthermore,
multi-objective optimization on integer variables is supported.
In contrast to \clingod{lp},
conditions on integer variables must be static.
% The system is available at \url{https://potassco.org/clingcon}.

Our systems are available at \url{https://potassco.org/labs/{clingoDL,clingoLP}}
and \url{https://potassco.org/clingcon}.

%%% Local Variables: 
%%% mode: latex
%%% TeX-master: "paper"
%%% End: 

\textbf{Big picture.}
Finally, let us relate our systems with others extending ASP with linear constraints.
The first category, referred to as translation-based approaches,
includes systems such as \ezsmt{}~\cite{liesus16a}, \dingo{}~\cite{jalini11a}, \aspmttosmt{}~\cite{barlee14b}, and \mingo{}~\cite{lijani12a}.
The first three translate both ASP and constraints into SAT Modulo Theories (SMT;~\cite{baseseti09a});
\dingo{} is restricted to difference constraints.
Unlike this,
\mingo's target formalism is Mixed Integer Linear Programming (MILP).
Furthermore, \aspartame~\cite{bageinospescsotawe15a} translates \ASPm{lc} (over integers) back to ASP by using the order encoding.
An advantage of translation-based approaches is that
once the input program is translated,
only a solver for the target formalism is needed.
In this way, they benefit from the features and performance of the respective target systems.
A drawback is the translation itself
since it may result in large propositional representations or weak propagation strength.
The second category extends the standard Conflict Driven Nogood Learning (CDNL;~\cite{gekasc09c}) machinery of ASP solvers with constraint propagators.
This allows for propagating both Boolean and linear constraints during search.
The latter are thus continuously checked for consistency and even new constraints may get derived.
For instance, the \clingo-based system \dlvhex[\textsc{cp}]~\cite{roeireri15a} uses \gecode, while \ezcsp{} uses a \prolog{} constraint solver
for consistency checking.
Unlike this,
\inca~\cite{drewal10a} extends a previous \clingo{} version with a customized lazy propagator generating constraints according to the order encoding.
This approach allows for deriving new constraints such as bounds of integer variables.

The \clingo{} derivatives \clingod{dl} and \clingod{lp} belong to the second category of systems, just like \clingcon~3.
%
% --------------------------------------------------------------------------------
\begin{table}[ht]
\caption{Comparing related applications}
\label{tab:related}
\center\small
\begin{threeparttable}
\begin{tabular}{@{\!\!}l@{\!\!}@{}c@{}@{}c@{}@{\!}c@{\!}@{\!}c@{\!}@{}c@{}@{\!}c@{\!}@{\!}c@{\!}@{\!}c@{\!}@{\!}c@{\!}@{\!}c@{\!}@{\!}c@{\!}}
\toprule
 & \clingo{}     & \clingo{}     & \clingcon{} & \aspartame{} & \inca{} & \ezcsp{} & \ezsmt{} & \mingo{} & \dingo{} & \sysfont{aspmt} & \dlvhex{}\\
 & [\textsc{dl}] & [\textsc{lp}] &             &              &         &          &          &          &          & \sysfont{2smt}  & [\textsc{cp}]\\\midrule
translation  & \xmark{} & \xmark{} & \cmark{}\tnote{1} & \cmark{} & \xmark{} & \xmark{} & \cmark{} & \cmark{} & \cmark{} & \cmark{} & \xmark{} \\
explicit     & \xmark{} & \xmark{} & \cmark{}\tnote{2} & \cmark{} & \cmark{} & \xmark{} & \xmark{} & \xmark{} & \xmark{} & \xmark{} & \xmark{} \\
non-linear   & \xmark{}\tnote{3} & \xmark{} & \cmark{}\tnote{4} & \cmark{}\tnote{4} & \cmark{} & \cmark{} & \cmark{} & \xmark{} & \xmark{}\tnote{3} & \cmark{} & \cmark{} \\
real numbers & \xmark{} & \cmark{} & \xmark{} & \xmark{} & \xmark{} & \cmark{} & \cmark{} & \cmark{}\tnote{5} & \xmark{} & \cmark{} & \xmark{}\\
optimization & \xmark{} & \cmark{}\tnote{6} & \cmark{} & \cmark{} & \cmark{} & \xmark{} & \xmark{} & \xmark{} & \xmark{} & \xmark{} & \cmark{} \\
non-tight    & \cmark{} & \cmark{} & \cmark{} & \cmark{} & \cmark{} & \cmark{} & \xmark{} & \cmark{} & \cmark{} & \xmark{} & \cmark{} 
\end{tabular}
\begin{tablenotes}\footnotesize
\begin{minipage}{0.5\textwidth}
\item[1] Allows for partial problem translations
\item[2] Lazily created
\item[3] Only difference constraints
\end{minipage}%
\begin{minipage}{0.45\textwidth}
\item[4] Translation of distinct into linear constraints
\item[5] Only for variables
\item[6] Optimization relative to stable models
\end{minipage}
\end{tablenotes}
\end{threeparttable}
\end{table}
%%% Local Variables:
%%% mode: latex
%%% TeX-master: "../paper"
%%% End:

% --------------------------------------------------------------------------------
%
Table~\ref{tab:related} summarizes important similarities and differences of the aforementioned systems.
The first row tells us whether a system relies on a translation to SMT, MILP, or ASP\@.
The second one indicates whether an approach uses some form of explicit variable representation.
This is the case when using an encoding and usually results in a large number of propositional atoms to represent variables with large domains.
Half of the systems are able to handle constraints over reals while the other half is restricted to integers.
Note that for a system of inequalities, a solution over reals can be found much easier than one over integers.
For all systems, real numbers are implemented as floating point numbers.
Due to this, round-off errors cannot completely be avoided.
Note that since computers are finite precision machines, 
the imprecision of floating point computations is common to any computer systems
and/or languages~\cite{goldberg91a}. 
\cplex\ uses numerically stable methods to perform its
linear algebra so that round-off errors usually do not cause problems.%
\footnote{See \emph{Numeric difficulties} at \url{https://www.ibm.com/support/knowledgecenter/SSSA5P_12.7.0/ilog.odms.studio.help/pdf/usrcplex.pdf}}
With ``non-linear'' we distinguish systems handling global or non-linear constraints,
and ``non-tight'' indicates whether a system can deal with recursive programs.
Finally, the table lists all systems that are able to optimize an objective function over integer and/or real variables.

%%% Local Variables:
%%% mode: latex
%%% TeX-master: "paper"
%%% End:

\section{Experimental analysis}\label{sec:experiments}

We begin with an empirical analysis of our \clingo{} derivatives in different settings.
We investigate, 
first,  different types of lc-atoms, viz.\ \lcatype{defined}{non-strict} versus \lcatype{external}{strict},
second, different levels of theory interfaces, \python\ or \cpp, for \clingod{dl},
and, third, different levels of integration, namely, dedicated implementations versus off-the-shelf solver.
Finally, we contrast the performance of our systems with other systems for \ASPm{lc}.

We ran each benchmark on a Xeon E5520 2.4 GHz processor under Linux limiting RAM to 20~GB and execution time to 1800s.
For \clingod{dl} and \clingod{lp}, we use \clingo~5.2.0.
Furthermore, we use \clingcon~3.2.0,
\dingo~v.2011-09-23,
\mingo~v.2012-09-30,
\ezsmt~1.0.0, 
and \ezcsp~1.7.9 for our experiments.
We upgraded \dingo\ and \mingo{} to use recent versions of their back-end solvers.
Hence, in our experiments,
the LP-based systems \clingod{lp} and \mingo{} use \cplex~12.7.0.0
and 
the SMT-based systems \dingo{} and \ezsmt{} use \zzz~4.4.2.
The benchmark set consists of 165 instances, 
among which 110 can be encoded using difference constraints~(\dl)
and 55 require linear constraints with more than two variables~(\lc).
In detail, the \dl\ set consists of
38 instances of two-dimensional strip packing~(\tsp)~\cite{sointabana10a},
and 72 instances of flow shop~(\fs), job shop~(\js), and open shop~(\os) problems~\cite{taillard93a},
selecting three instances for each job and machine at random.
Since not all systems support optimization over variable values,
we bounded the instances with 1.2 times the best known bound and solved the resulting decision problem.
The \lc\ instance set includes
20 instances of incremental scheduling~(\is),
15 instances of reverse folding~(\rf), and
20 instances of weighted sequence~(\ws).
Encodings have been adopted from ~\cite{liesus16a}
in combination with the instances from the ASP competition.%
\footnote{We refrained from using the other three benchmark classes from this source because the available instances were too easy in view of producing informative results.}
Our empirical evaluation focuses on available systems sharing comparable encodings.
This was not the case for \aspartame, \aspmttosmt, \inca, and \dlvhex[\textsc{cp}].
The first two systems have a proper and thus different input language and encoding philosophy,
\inca\ produced incorrect results (cf.~\cite{bakaossc16a} for details),
and \dlvhex[\textsc{cp}] is no longer maintained.

Table~\ref{tab:clingolc} compares \clingod{dl} and \clingod{lp} with different 
encoding techniques, 
types of theory atoms, and 
programming language hosting the theory interface
by measuring average time~(\T) and timeouts~(\TO).
% --------------------------------------------------------------------------------
\begin{table}%
\newcommand{\mc}[3]{\multicolumn{#1}{#2}{#3}}%
\caption{Comparison of \clingo\ derivatives \clingod{dl} and \clingod{lp}\label{tab:clingolc}}%
\centering%
\begin{tabular}{@{}r@{}@{}r@{}@{}r@{}@{}r@{\ }r@{}@{}r@{\ }r@{}@{}r@{\ }r@{}@{}r@{\ }r@{}@{}r@{\ }r@{}@{}r@{\ }}
\toprule
       &        & \mc{2}{@{\,}c@{\,}}{\clingolc{dl}{dns}{py}} 
                & \mc{2}{@{\,}c@{\,}}{\clingolc{dl}{es}{py}} 
                & \mc{2}{@{\,}c@{\,}}{\clingolc{lp}{dns}{py}} 
                & \mc{2}{@{\,}c@{\,}}{\clingolc{lp}{es}{py}} 
                & \mc{2}{@{\,}c@{\,}}{\clingolc{dl}{dns}{cpp}} 
                & \mc{2}{@{\,}c@{\,}}{\clingolc{dl}{es}{cpp}}  \\
\class & \#inst &   \T &       \TO  &   \T &       \TO  &   \T & \TO &   \T & \TO &           \T &         \TO &         \T &       \TO  \\
\midrule
\tsp   &     38 &  344 &         6  &  484 &         9  & 1346 &  23 & 1753 &  36 & \textbf{148} &  \textbf{3} &       342  &         7  \\
\fs    &     35 &  678 &        11  & 1541 &        27  & 1221 &  21 & 1800 &  35 & \textbf{465} &  \textbf{5} &      1349  &        26  \\
\js    &     24 & 1261 &        15  & 1229 &        14  & 1800 &  24 & 1800 &  24 & \textbf{534} &  \textbf{4} &       678  &         7  \\
\os    &     13 &    8 & \textbf{0} &   17 & \textbf{0} &  963 &   6 & 1532 &  10 &   \textbf{0} &  \textbf{0} & \textbf{0} & \textbf{0} \\
\midrule
\dl    &    110 &  611 &        32  &  928 &        50  & 1360 &  74 & 1752 & 105 & \textbf{316} & \textbf{12} &       695  &        40  \\
\bottomrule
\end{tabular}%
\end{table}%
%
%%% Local Variables:
%%% mode: latex
%%% TeX-master: "../paper"
%%% End:

% --------------------------------------------------------------------------------
Each column consists of one combination of form \emph{system}/\emph{atom}/\emph{language}, 
where
\emph{system} is either \textsc{dl} or \textsc{lp} for \clingod{dl} and \clingod{lp},
\emph{atom} either \atomtype{dns} or \atomtype{es} for \lcatype{defined}{non-strict} and \lcatype{external}{strict} lc-atoms,
and
\emph{language} either \textsc{py} or \textsc{cpp} for \python\ and \cpp, respectively.
To compare \clingod{dl} and \clingod{lp}, we restrict the set of benchmarks to \dl.
We observe that \atomtype{dns} performs better than \atomtype{es} in all settings.
Under lc-stable model semantics, defined lc-atoms are more tightly constrained.
External lc-atoms, on the other hand, induce an implicit choice leading to a
larger search space and might introduce duplicate solutions with different assignments.
Furthermore, strict lc-atoms double the amount of implications 
that have to be considered by the propagator.
As expected, the \cpp\ variant of \clingod{dl} outperforms its \python\ counterpart,
even though the performance gain does not reach an order of magnitude.

Table~\ref{tab:systems} compares different systems dealing with \ASPm{lc}
by average time~(\T) and timeouts~(\TO).
%
% --------------------------------------------------------------------------------
\begin{table}%
\newcommand{\mc}[3]{\multicolumn{#1}{#2}{#3}}%
\caption{Comparison of different systems for ASP with linear constraints\label{tab:systems}}%
\centering%
\begin{tabular}{@{}r@{\!}@{\!}r@{\!}
  @{\ }r@{\hspace{-5pt}}@{\hspace{-5pt}}r@{\,}
  @{\ }r@{\hspace{-5pt}}@{\hspace{-5pt}}r@{\,}
  @{\ }r@{\hspace{-1pt}}@{\hspace{-1pt}}r@{\,}
  @{\ }r@{\hspace{-1pt}}@{\hspace{-1pt}}r@{\,}
  @{\ }r@{\hspace{-1pt}}@{\hspace{-1pt}}r@{\,}
  @{\ }r@{\hspace{-1pt}}@{\hspace{-1pt}}r@{\,}
  @{\ }r@{\hspace{-1pt}}@{\hspace{-1pt}}r@{}}
\toprule
       &        & \mc{2}{@{}c@{}}{\clingolc{dl}{dns}{cpp}} 
                & \mc{2}{@{}c@{}}{\clingolc{lp}{dns}{py}} 
                & \mc{2}{@{}c@{}}{\clingcon} 
                & \mc{2}{@{}c@{}}{\dingo} 
                & \mc{2}{@{}c@{}}{\mingo} 
                & \mc{2}{@{}c@{}}{\ezsmt} 
                & \mc{2}{@{}c@{}}{\ezcsp}                  \\
\class & \#inst &          \T  &        \TO  &   \T & \TO & \T & \TO & \T & \TO & \T & \TO & \T & \TO & \T & \TO\\
\midrule
\tsp   &    38  &         148  &          3  & 1346 & 23 & \textbf{3}    & \textbf{0}  & 403  & 7  & 292  & 5  & 318  & 6  & 1800 & 38\\
\fs    &    35  & \textbf{465} &  \textbf{5} & 1221 & 21 & 1022 & 19 & 1047 & 20 & 1040 & 16 & 1667 & 32 & 735  & 9\\
\js    &    24  &         534  &          4  & 1800 & 24 & \textbf{277}  & \textbf{3}  & 1258 & 15 & 1423 & 18 & 1315 & 15 & 1800 & 24 \\
\os    &    13  &   \textbf{0} &  \textbf{0} &  963 & 6  & 1    & \textbf{0}  & 4    & \textbf{0}  & 76   & \textbf{0}  & 24   & \textbf{0}  & 1044 & 7 \\
\midrule
\dl    &    110 & \textbf{316} & \textbf{12} & 1360 & 74 & 387  & 22 & 765  & 42 & 743  & 39 & 930  & 52 & 1372 & 78 \\
\midrule
\is    &    20  &         \na  &         \na & 1800 & 20 & \textbf{582}  & \textbf{5}  & \na & \na & 649  & 7  & 648  & 7  & 1620 & 18\\
\rf    &    15  &         \na  &         \na & 1680 & 14 & \textbf{21}   & \textbf{0}  & \na & \na & 542 & 1 & 121  & \textbf{0}  & 1013 & 7\\
\ws    &    20  &         \na  &         \na & 1800 & 20 & 27   & \textbf{0}  & \na  & \na & 90   & \textbf{0}  & \textbf{12}   & \textbf{0}  & 1800 & 20 \\
\midrule
\lc    &    55  &         \na  &         \na & 1767 & 54 & \textbf{227}  & \textbf{5}  & \na  & \na & 416  & 8 & 273  & 7  & 1520 & 45\\
\midrule
all    &    165 &         \na  &         \na & 1564 & 128 & \textbf{307}  & \textbf{27} & \na  & \na & 580  & 47 & 602  & 59 & 1446 & 123\\
\bottomrule
\end{tabular}
\end{table}
%
%%% Local Variables:
%%% mode: latex
%%% TeX-master: "../paper"
%%% End:

% --------------------------------------------------------------------------------
%
Only the best configurations from Table~\ref{tab:clingolc} were selected for comparison.
All systems were tested using their default configurations.
For \dl{}, \clingolc{dl}{dns}{cpp} performs best overall,
even though \clingcon\ is better for \tsp\ and \js.
The class \fs\ generates the most difference constraints among all benchmark classes,
making it less suited for translation-based approaches, like \dingo, \mingo, and \ezsmt,
and producing overhead for more involved propagation as in \clingcon.
By default, \ezcsp\ performs the theory consistency check on full answer sets,
and by doing so avoids handling vast amounts of constraints during search
and therefore performs comparatively well on \fs.
For the other classes though, this generate and test approach is less effective.
Regarding \lc\ and overall results, \clingcon\ clearly dominates the competition,
followed by the two translation-based approaches \mingo\ and \ezsmt\ 
with underlying state-of-the-art solvers \cplex\ and \zzz, respectively.
\clingolc{lp}{dns}{py} falls behind,
since it is a straightforward \python\ implementation
and uses an exponential consistency check.
In addition, distinct features of \clingod{lp} like real-valued variables and optimization
as well as dynamic conditions are not supported by other systems 
and thus not included in the benchmark set.

%%% Local Variables: 
%%% mode: latex
%%% TeX-master: "paper"
%%% End: 

\section{Summary}\label{sec:summary}

We presented several truly hybrid ASP systems incorporating difference and linear constraints.
Previous approaches addressed this by resorting to translations into foreign solving paradigms like MILP or SMT.
This difference is analogous to the one between genuine ASP solvers like \clasp{} and \wasp{}
and earlier ones like \assat\ and \cmodels\ that translate ASP to SAT.
The resulting systems \clingod{dl} and \clingod{lp} comprise several complementary aspects.
For instance, \clingod{dl} relies upon customized propagators, one variant using a \python{} API, the other a \cpp{} API.
This is similar to the approach of \inca{} and \clingcon~3 for Constraint ASP.
Unlike this, \clingod{lp} builds upon the \python{} API to incorporate off-the-shelf LP solvers for propagation, optionally \cplex{} or \lpsolve.
This is similar to the approach of \dlvhex[\textsc{cp}] and \clingcon~2 integrating \gecode{} for constraint processing.
Both \clingod{dl} and \clingod{lp} allow for dealing with integer as well as real variables.
The former admits two, the latter an arbitrary number of such variables per linear constraint.
This is complemented by \clingcon~3 adding constraint processing to \clingo{} by using a low level API.

We accomplished this by instantiating the generic framework of ASP modulo theories described in~\cite{gekakaosscwa16a}.
We defined lc-stable models and elaborated upon different types of lc-atoms,
ultimately settling on the combinations \lcatype{defined}{non-strict} and \lcatype{external}{strict}
for \clingod{dl} and \clingod{lp}.%
\footnote{This is our recommendation in view of our analysis in Section~\ref{sec:background};
  both systems actually support all four combinations of strict/non-strict and defined/external lc-atoms.}
Our underlying formal analysis on the interaction of strict- and definedness has actually a much broader impact
given that other systems follow similar principles.
Although we lack a deeper analysis,
\inca{} and \dlvhex[\textsc{cp}] appear to adhere to an \lcatype{external}{strict} treatment of constraint atoms,
just as our previous systems \clingcon, \dingo, and \mingo, 
while \ezsmt{} and \ezcsp{} seem to follow an \lcatype{external}{non-strict} approach.
Moreover, the results in Proposition~\ref{thm:settings} are of a general nature and apply well beyond ASP systems dealing with linear constraints.

We provided a conceptual and empirical comparison of \clingod{dl} and \clingod{lp} with related systems 
for dealing with different forms of linear constraints in ASP.
Our experiments focused on,
first, examining different types of lc-atoms and APIs for both \clingo{} derivatives,
and, second, comparing them with related systems.
In the first case, \clingod{dl} using \lcatype{defined}{non-strict} lc-atoms along with the \cpp\ API 
yields the best results,
and in the second one, 
the aforementioned \clingod{dl} configuration outperforms the other systems for the set of benchmarks only involving difference constraints,
and \clingcon\ has an edge over all other systems regarding the set of benchmarks featuring arbitrary (integer-based) linear constraints.

Finally, % Last but not least,
we showed how easily our machinery can be applied to online reasoning scenarios
by using \clingo's multi-shot and theory reasoning capabilities in tandem.
% and illustrated this by encoding the spoiled Yale shooting.

%%% Local Variables: 
%%% mode: latex
%%% TeX-master: "paper"
%%% End: 

\bibliographystyle{acmtrans}
% \bibliography{lit,procs,akku,own}

%%% Local Variables: 
%%% mode: latex
%%% TeX-master: "paper"
%%% End: 

% \input{appendix}
\end{document}